\documentclass[10pt,twocolumn,letterpaper]{article}

\pdfoutput=1

\usepackage{iccv}
\usepackage{times}
\usepackage{epsfig}
\usepackage{graphicx}
\usepackage{amsmath}
\usepackage{amssymb}

\usepackage{subcaption}
\usepackage{multirow}
\usepackage{hhline}
\usepackage{bbm}
\usepackage{tabularx}
\usepackage{threeparttable}
\usepackage[boxed]{algorithm2e}

\newcommand{\epsintra}{$eps_\mathrm{intra}$}
\newcommand{\epsinter}{$eps_\mathrm{inter}$}

\usepackage[breaklinks=true,bookmarks=false]{hyperref}

\iccvfinalcopy 

\def\iccvPaperID{6939} 

\ificcvfinal\fi

\begin{document}

\title{A Free Lunch to Person Re-identification: Learning from Automatically Generated Noisy Tracklets}

\author{\textbf{Hehan Teng}\textsuperscript{1}, \textbf{Tao He}\textsuperscript{1}, \textbf{Yuchen Guo}\textsuperscript{1}, \textbf{Zhenhua Guo}\textsuperscript{2}, \textbf{Guiguang Ding}\textsuperscript{1}\\
\textsuperscript{1} School of Software, Tsinghua University, Beijing, China\\
\textsuperscript{2} Alibaba Group\\
{\small thss15\_tenghh@163.com, \{kevin.92.he, yuchen.w.guo, cszguo\}@gmail.com, dinggg@tsinghua.edu.cn}
}

\maketitle
\ificcvfinal\thispagestyle{empty}\fi

\begin{abstract}
A series of unsupervised video-based re-identification (re-ID) methods have been proposed to solve the problem of high labor cost required to annotate re-ID datasets. But their performance is still far lower than the supervised counterparts. In the mean time, clean datasets without noise are used in these methods, which is not realistic. In this paper, we propose to tackle this problem by learning re-ID models from automatically generated person tracklets by multiple objects tracking (MOT) algorithm.
To this end, we design a tracklet-based multi-level clustering (TMC) framework to effectively learn the re-ID model from the noisy person tracklets. First, intra-tracklet isolation to reduce ID switch noise within tracklets; second, alternates between using inter-tracklet association to eliminate ID fragmentation noise and network training using the pseudo label. Extensive experiments on MARS with various manually generated noises show the effectiveness of the proposed framework. Specifically, the proposed framework achieved mAP 53.4\% and rank-1 63.7\% on the simulated tracklets with strongest noise, even outperforming the best existing method on clean tracklets. Based on the results, we believe that building re-ID models from automatically generated noisy tracklets is a reasonable approach and will also be an important way to make re-ID models feasible in real-world applications.
\end{abstract}


\section{Introduction}

Person re-identification (re-ID) is to match persons across non-overlapping cameras. It is one of the core techniques in intelligent surveillance analysis. Due to the urgent demand for public safety, it has been an active research field over the years. Video-based person re-ID is the problem where subjects to be retrieved are presented as video sequences. Person re-ID has shown promising results in a fully supervised setting. This learning paradigm assumes that there is a large number of labeled high-quality cross-camera training data. But it is of the high cost to collect such a large-scale dataset, due to the exponential labeling cost. Besides, a well-trained re-ID model has been proved to perform much worse in a new domain. 

\begin{figure*}
    \centering
    \includegraphics[width=.8\linewidth]{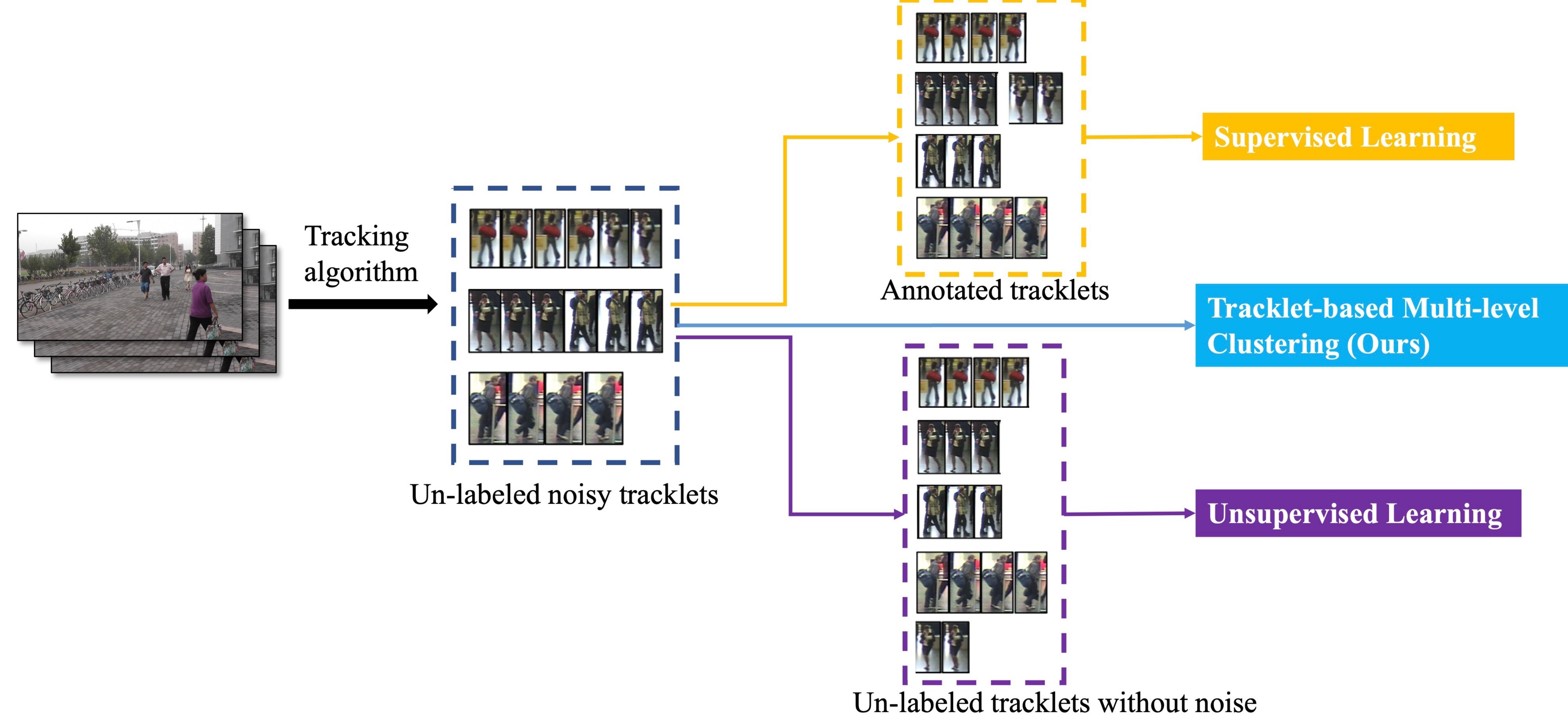}
    \caption{Different categories of person re-ID methods. Supervised methods require full annotation of video sequences. Datasets used by unsupervised video re-ID methods require human labor to eliminate noise within tracklets generated by MOT algorithms. Our method uses MOT tracklets directly.}
    \label{fig:method-differences}
\end{figure*}

To overcome the drawbacks of supervised methods, in the last two years, several works have turned to study unsupervised or weakly supervised person re-ID. Specifically, we focus on unsupervised video-based person re-ID, where training data can be obtained without human labor by multiple object tracking~\cite{ciaparrone2020mot} (MOT) algorithms, as shown in figure~\ref{fig:method-differences}. Most of the existing unsupervised video re-ID methods still yield unsatisfactory results. Moreover, these methods operate on video re-ID datasets, such as MARS~\cite{zheng2016mars}, iLIDS-VID~\cite{wang2014ilids} and PRID 2011~\cite{hirzer2011prid}. It should be noted that, as shown in figure~\ref{fig:method-differences}, although these datasets are used in a unsupervised manner, i.e. video sequences without intra- and inter-camera ID association, the sequences themselves are clean and without noise, and the production of such clean sequences requires substantial human effort as well. The gap between such datasets and MOT-generated tracklets are the noise introduced by MOT algorithms, which is mainly ID fragmentation noise and ID switch noise, alongside with detection~\cite{ren2016detecion} noise. Some methods have been proposed to exploit tracklets to build a re-ID model, but only ID fragmentation noise is considered, while ID switch and detection noise are ignored, resulting in a wider gap from being applicable to real-life scenarios.

We propose a new tracklet-based clustering and fine-tuning framework to account for both ID fragmentation noise and ID switch noise in MOT-generated tracklets. By analyzing characteristics of aforementioned noise, a multi-stage clustering-based method is proposed to reduce noise in the tracklets before feeding them into the unsupervised training pipeline, resulting in significant performance boost. Since raw video of video re-ID datasets is generally unavailable, a novel algorithm is proposed to generate simulated tracklets from video re-ID datasets, to assist evaluating our method under various strength of noise.

We summarize our contribution as three-fold. 

\begin{enumerate}
    \item Firstly, we propose taking raw tracklets generated by MOT algorithms as input of our method, removing the requirement of human effort completely, resulting in nearly zero-cost training data preparation, moving a step closer into solving realistic problems. 
    \item Secondly, we analyze dominant noise categories in tracklets generated by MOT algorithms, i.e. ID fragmentation and ID switch noise, revealing characteristics which is exploited in our novel noise reduction processing. We combine the noise reduction techniques with a self-training mechanism in our method, named Tracklet-based Multi-level Clustering (TMC).
    \item Thirdly, experiments show that our method achieves remarkable performance given that realistic tracklets with noise are used (mAP 55.3\% rank-1 68.2\% on simulated tracklets), and, if clean video re-ID datasets are used instead, outperforms existing unsupervised video re-ID methods.
\end{enumerate}

\section{Related Work}

\subsection{Unsupervised video-based re-ID }

Although unsupervised video-based person re-ID is a relatively unexplored area compared to the supervised problem, it is gaining attention over the past few years. In this setting, unlabeled video sequences (tracklets) without noise are provided, and each person may have more than one tracklets under a certain camera. From MOT point of view, within each camera, these tracklets have ID fragmentation noise, but not ID switch noise.
Dynamic Graph Matching (DGM)~\cite{ye2017dgm} leverages the graph matching technique by constructing a graph for samples in each camera for label estimation, and iteratively update the graph to produce estimated label. To further improve the performance, global camera network constraints~\cite{ye2019constraint} are exploited for consistent matching. 
Riachy \etal~\cite{riachy2019riachy} formulates the re-ID task as a set-based matching problem that they tackle using the NBNN classifier.
DAL~\cite{chen2018dal} proposed by Chen\etal is an end-to-end deep learning method which jointly optimize two association losses.
EUG~\cite{wu2018eug} is a step-wise one-shot learning method, gradually selecting a few candidates from unlabeled tracklets to enrich the labeled tracklet set.
RACE~\cite{ye2018race} is robust anchor embedding method iteratively assigns labels to the unlabelled tracklets to enlarge the anchor video sequences set. Note that the EUG and RACE, requires additional information to initialize their learning process, which usually involves extra human labor.

\begin{figure*}
\centering
    \begin{subfigure}[t]{0.9\textwidth}
        \centering
        \includegraphics[width=\linewidth]{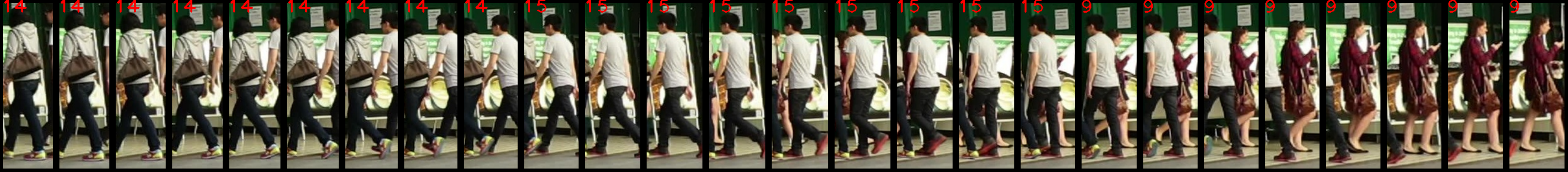}
        \caption{ID Switch}
    \end{subfigure}
    
    \begin{subfigure}[t]{0.9\textwidth}
    \begin{subfigure}[t]{0.5\textwidth}
        \centering
        \includegraphics[width=\linewidth]{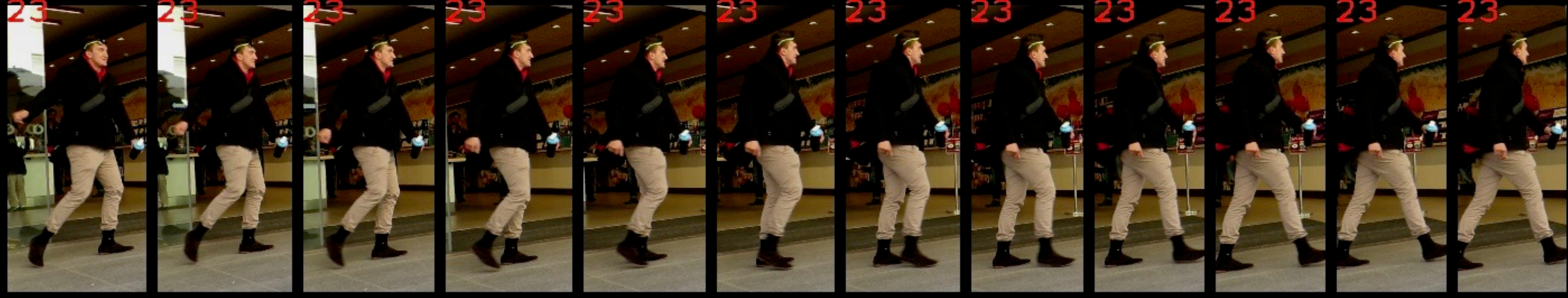}
    \end{subfigure}
    \hfill
    \begin{subfigure}[t]{0.48\textwidth}
        \centering
        \includegraphics[width=\linewidth]{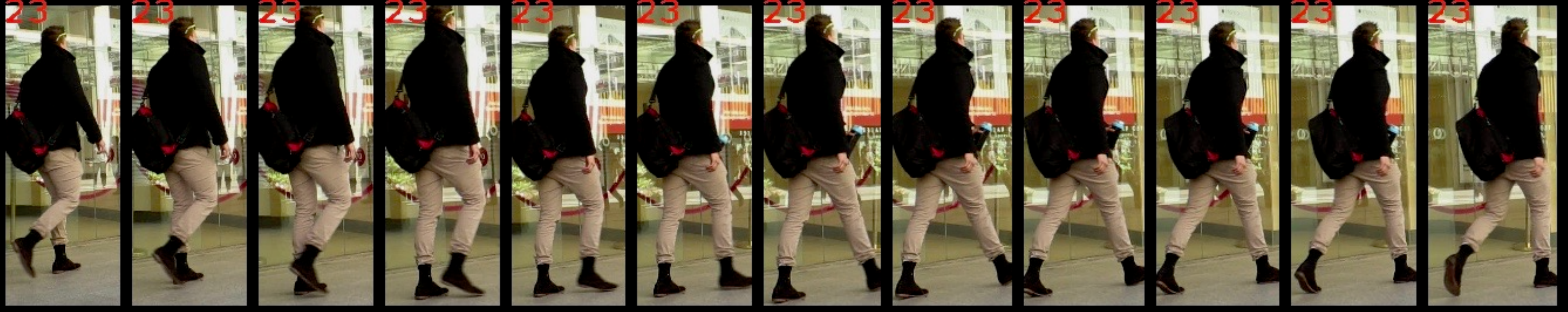}
    \end{subfigure}
    \centering
    \caption{ID Fragmentation}
    \end{subfigure}
    
    \centering
    \caption{Examples of noise types in tracklets. (a) Example of ID switch. The tracklet each contains more than one (ground-truth) person. (b) Example of ID fragmentation. The occurrence of a (ground-truth) person was split up in more than one tracklet. Red numbers in the top-left corner of each frame denotes the ground-truth ID of the person tracked in that frame.}
    \label{fig:noisy_tracklets}
\end{figure*}

\subsection{Tracklet-based re-ID }

Methods based on tracklets are developed to be independent of source data or model and also save human efforts. UTAL~\cite{li2019utal} is the first work to build a re-ID model from tracklets, which are automatically generated by multiple object tracking (MOT) algorithms. But only fragmentation noise is considered in this work, and the identity switch and detection noises, which commonly occur in the generated tracklets, are ignored in this work. Furthermore, the experiments on two commonly used datasets, in fact, are particularly designed to only contain the inter-camera fragmentation noise, without intra-camera noise. Their performance is also much lower than the domain adaptation or supervised counterparts. TSSL~\cite{wu2020tssl} is also built on tracklets but without camera view information. The experimental results on the Market are similarly low as UTAL. There are other works that require human efforts to label part of the dataset. MTML~\cite{zhu2019mtml} assumes that the intra-camera instances are perfectly labeled, and it performs multi-task learning on multiple camera views. The progressive framework by Wu \etal~\cite{wu2019wu} also requires the label of one example for each pedestrian. Although involved human efforts, the performance of these methods is still far behind the supervised ones, which hinders the application in the real world.

\section{Method}

\subsection{Tracklet Noise Analysis}
We present the definition of noise types in tracklets before analyzing their characteristics. In the evaluation metrics used in MOT benchmarks, such as MOT20~\cite{dendorfer2020mot20}, the determination of ID fragmentation and ID switch requires a global tracklet-to-ground-truth assignment produced by optimal matching using Hungarian algorithm~\cite{kuhn1955hungarian}. In this work we use a simplified and more intuitive definition as follows. ID fragmentation is said to occur if the occurrence of a (ground-truth) person was split up in more than one tracklet, and ID switch is said to occur if a tracklet contains more than one (ground-truth) person. Figure~\ref{fig:noisy_tracklets} shows examples for both types of noise.

MOT-generated tracklets cannot be used to train re-ID model directly since both types of noise have a negative impact on the learning process. ID fragmentation causes loss of information since during the learning process, multiple tracklets of a person would have been treated as different IDs. A supervised method will try to guide the model to differentiate these IDs while they are actually the same person, effectively misleading the model. On the other hand, an unsupervised method would have to discover and associate multiple tracklets of a person. Tracklets with ID switch have internal inconsistency, which affect both supervised and unsupervised method in the same way. These tracklets guide the system into treating relevant people as same, and will harm the ability the model to distinguish different people.

\subsection{Tracklet-based Multi-level Clustering}

A multi-level clustering method is proposed to reduce both ID fragmentation and ID switch noise in the tracklets. A low-level clustering, named ``intra-tracklet isolation'' is first done within each tracklet, forming groups of images, to reduce ID switch noise. A high-level clustering, named ``inter-tracklet association'' is then conducted to mine association between tracklets to reduce ID fragmentation noise.

\begin{figure}
  \begin{subfigure}[t]{.23\textwidth}
    \centering
    \includegraphics[width=\linewidth]{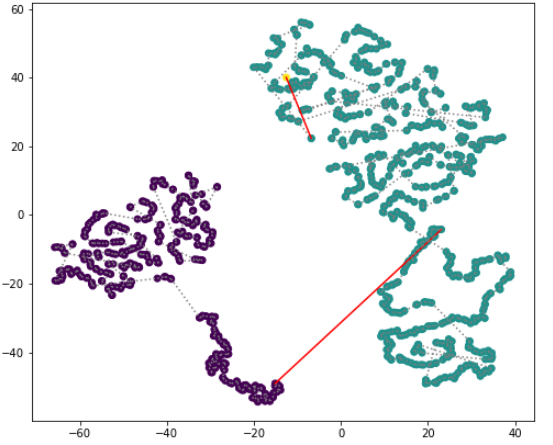}
    \caption{}
  \end{subfigure}
  \hfill
  \begin{subfigure}[t]{.24\textwidth}
    \centering
    \includegraphics[width=\linewidth]{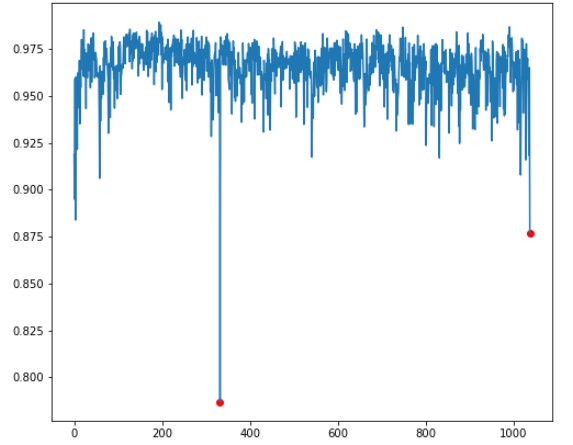}
    \caption{}
  \end{subfigure}

  \medskip

  \begin{subfigure}[t]{.23\textwidth}
    \centering
    \includegraphics[width=\linewidth]{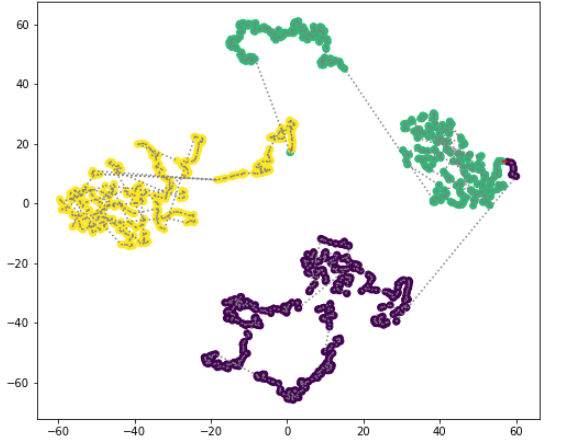}
    \caption{}
  \end{subfigure}
  \hfill
  \begin{subfigure}[t]{.24\textwidth}
    \centering
    \includegraphics[width=\linewidth]{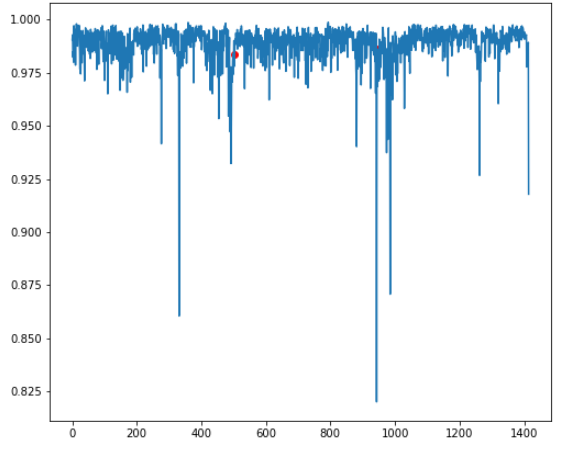}
    \caption{}
  \end{subfigure}
  
  \caption{Illustration of tracklets with ID switch noise. Figures (a)(c) show the feature distribution of images in the tracklets, points colored according to ground-truth person ID. Figures (b)(d) show the cosine similarity of adjacent frames in the tracklet. Figures (a)(b), (c)(d) are from the same tracklet, respectively.}
  \label{fig:tsne-similarity}
\end{figure}

\begin{figure}
  \begin{subfigure}[t]{.23\textwidth}
    \centering
    \includegraphics[width=\linewidth]{tsne-1.pdf}
    \caption{}
  \end{subfigure}
  \hfill
  \begin{subfigure}[t]{.23\textwidth}
    \centering
    \includegraphics[width=\linewidth]{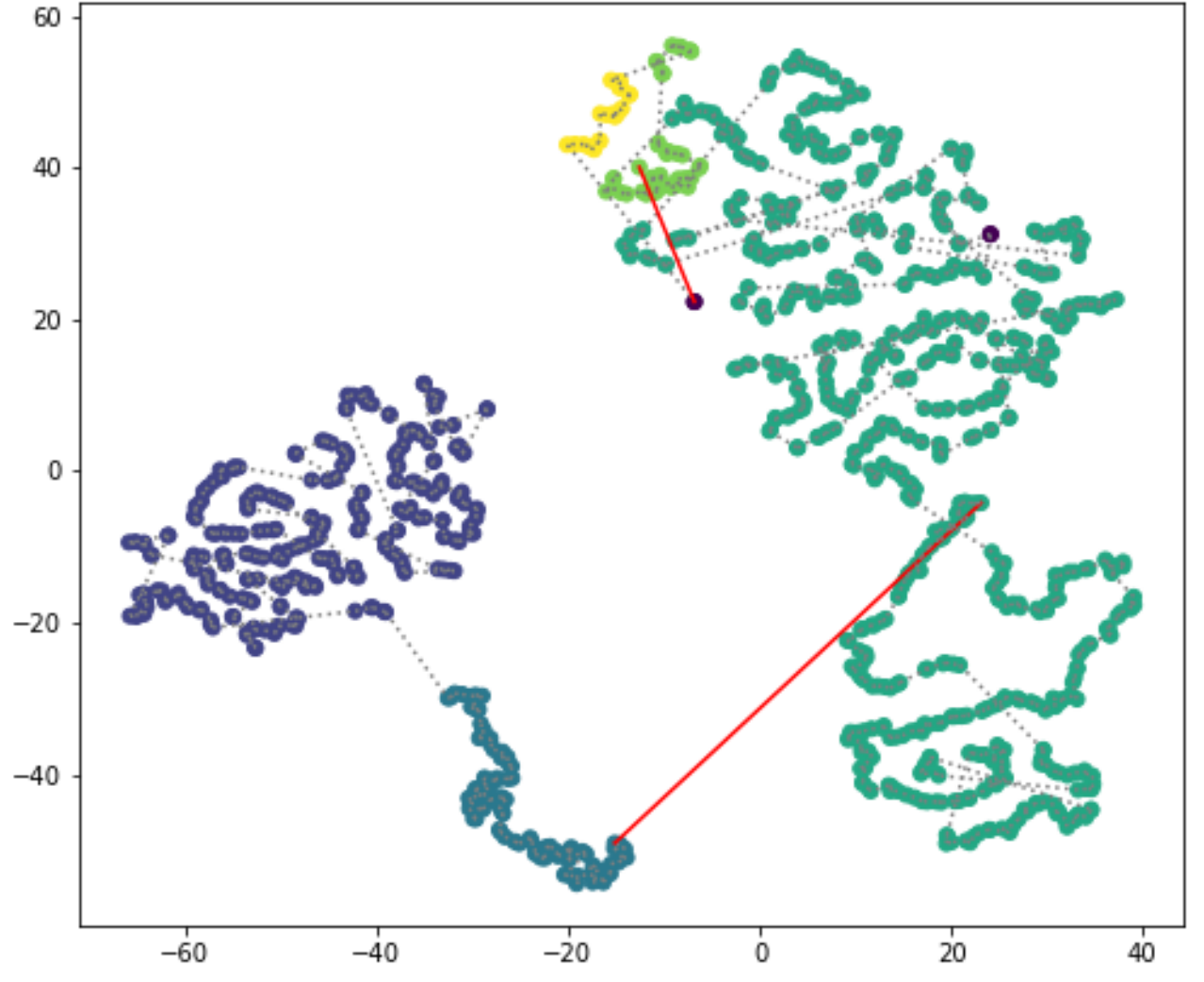}
    \caption{}
  \end{subfigure}

  \medskip

  \begin{subfigure}[t]{.23\textwidth}
    \centering
    \includegraphics[width=\linewidth]{tsne-2.pdf}
    \caption{}
  \end{subfigure}
  \hfill
  \begin{subfigure}[t]{.23\textwidth}
    \centering
    \includegraphics[width=\linewidth]{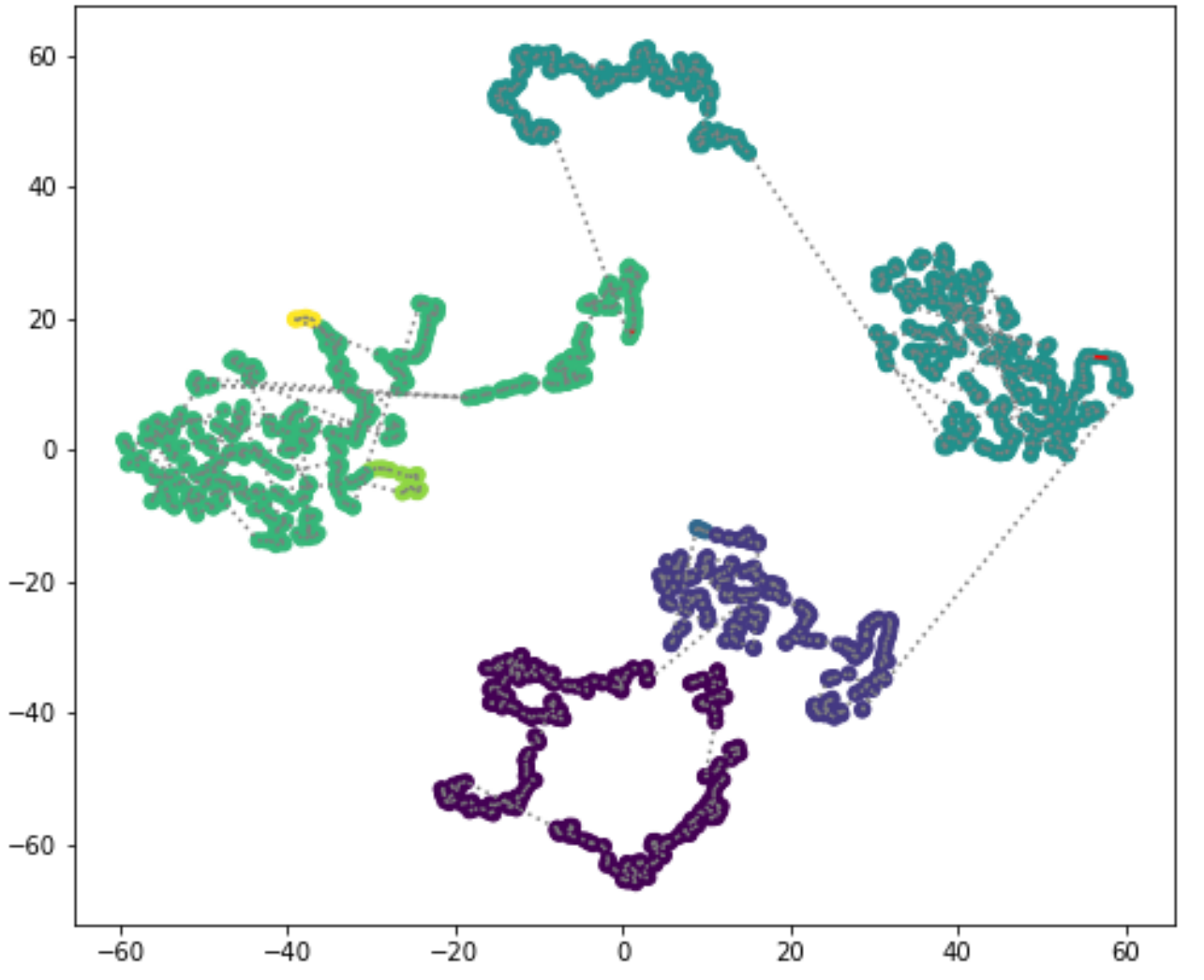}
    \caption{}
  \end{subfigure}
  
  \caption{Results of intra-tracklet clustering. Figures show the feature distribution of images in the tracklets with ID switch. Points in figures (a)(c) colored according to ground-truth person ID. Points in figures (b)(d) colored according to clustering results.}
  \label{fig:intra-cluster-vis}
\end{figure}

\begin{figure}
    \centering
    \includegraphics[width=.8\linewidth]{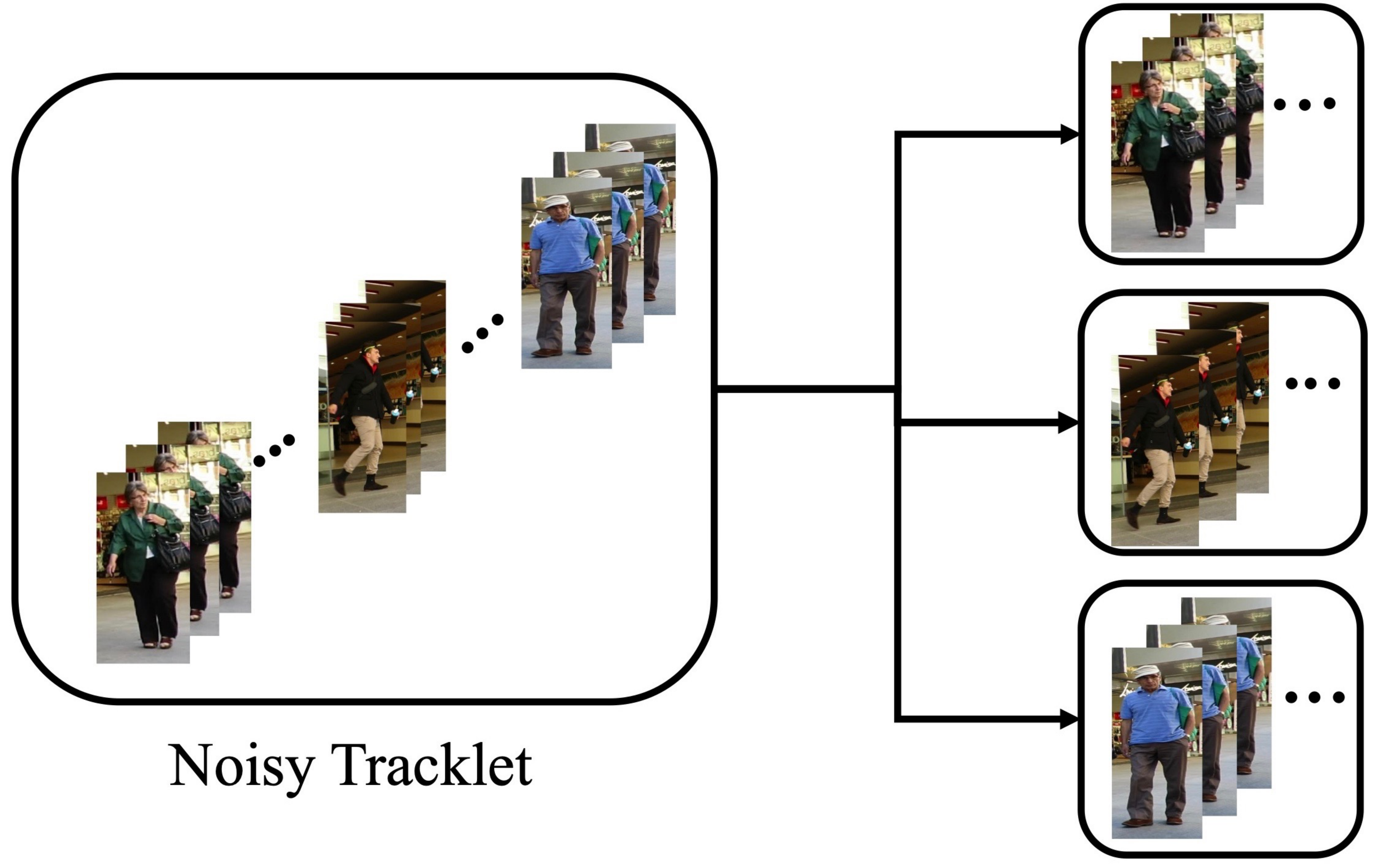}
    \caption{Intra-tracklet isolation. Tracklets are clustered into smaller but cleaner tracklets. Images from each resulting cluster will be composed into a smaller tracklet, while preserving the chronological order.}
    \label{fig:intra-tracklet-cluster}
\end{figure}

\noindent \textbf{Intra-tracklet isolation }
We propose an intra-tracklet isolation step to reduce ID switch noise in MOT-generated tracklets. Features of images in the tracklets are extracted by ImageNet-pretrained~\cite{deng2009imagenet} ResNet50 and visualized with t-SNE~\cite{van2008tsne} in figures~\ref{fig:tsne-similarity} (a)(c). It is shown that features of different person in a tracklet are roughly distinguishable in the feature space, indicating that occurrences of ID switch do not result from similarities of people appearance, instead, are more of a result of indistinguishable movement trajectories, which is consistent with what is observed in real-life MOT algorithm results. Figures~\ref{fig:tsne-similarity} (b)(d) show cosine similarity of features between adjacent frames in the tracklet. It can be shown in figure~\ref{fig:tsne-similarity} (b) that similarity is significantly low when ID switch occurs (red points). This indicator, however, is unreliable as illustrated in figure~\ref{fig:tsne-similarity} (d).

Therefore, intra-tracklet isolation was used to process tracklets into smaller but cleaner tracklets (figure~\ref{fig:intra-tracklet-cluster}). Specifically, DBSCAN~\cite{ester1996dbscan} was used to cluster the features. Images from each resulting cluster will be composed into a smaller tracklet, while preserving the chronological order. Figure~\ref{fig:intra-cluster-vis} shows the result of intra-tracklet isolation.

\noindent \textbf{Inter-tracklet association }
While intra-tracklet isolation reduces ID switch noise, it introduces extra ID fragmentation noise, in addition to inherent ID fragmentation noise in tracking results. We propose using inter-tracklet association step to mine association of tracklets from the same person. Consecutive images of fixed length were sampled for each tracklet. Features of images in a tracklet are extracted using the updated model, followed by an average-pooling step to produce the feature of the tracklet. Features of all tracklets are then clustered with DBSCAN to produce hard pseudo labels.

\begin{figure*}
    \centering
    \includegraphics[width=.8\linewidth]{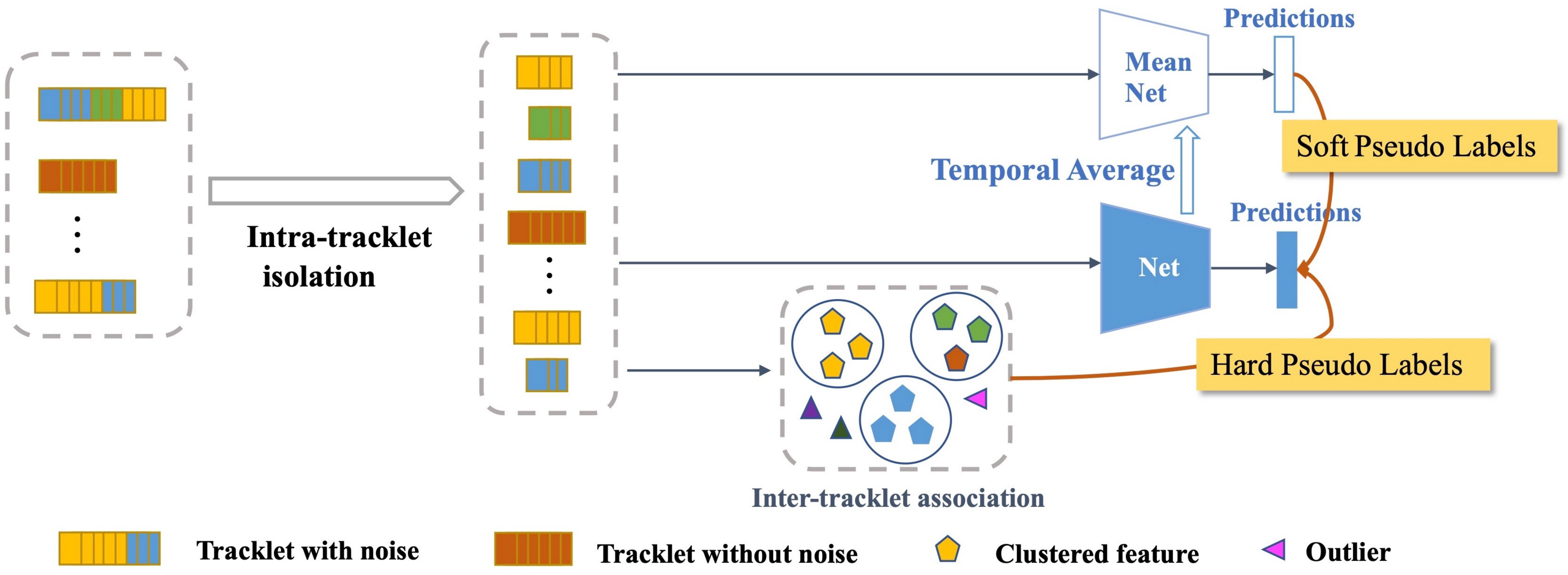}
    \caption{The overall pipeline. The tracklets with noise are first processed with intra-tracklet isolation to produce a new collection of tracklets with less noise. At the start of each training epoch, hard pseudo label are generated using inter-tracklet association. In each iteration of a epoch, the same input images as Net were feed into Mean Net to produce soft pseudo labels. The Net is trained by utilizing both hard pseudo labels and soft pseudo labels. Updated Net is then used to update Mean Net in EMA (Exponential Moving Average)~\cite{klinker2011ema} mode.}
    \label{fig:framework}
\end{figure*}

\subsection{Training pipeline }

A simplification of MMT~\cite{ge2019mmt}~\cite{zhang2018mutual} which keeps only one network (``student'') and its past temporal average model (``teacher'') was adopted as our overall training pipeline. The former is denoted as ``Net'' and latter as ``Mean Net''~\cite{tarvainen2017meanteacher} in figure~\ref{fig:framework}. 

\noindent \textbf{Framework } 
The tracklets with noise are first processed with intra-tracklet isolation to produce a new collection of tracklets with less noise. At the start of each training epoch, hard pseudo label are generated using inter-tracklet association as described above. In each iteration of an epoch, the same input images as Net were fed into Mean Net to produce soft pseudo labels. The Net is trained by utilizing both hard pseudo labels and soft pseudo labels. Updated Net is then used to update Mean Net in EMA (Exponential Moving Average) mode.

\begin{equation}\label{eq:ema}
E^{(T)}\left[\boldsymbol{\theta}\right]=\alpha E^{(T-1)}\left[\boldsymbol{\theta}\right]+(1-\alpha) \boldsymbol{\theta}
\end{equation}

where $E^{(T-1)}\left[\boldsymbol{\theta}\right]$ indicate the temporal average parameters of Net, i.e. Mean Net in the previous iteration, following the notation in \cite{ge2019mmt}. The initial Mean Net parameters  $E^{(0)}\left[\boldsymbol{\theta}\right] = \boldsymbol{\theta}$, and $\alpha$ is the ensembling momentum to be within the range $[0,1)$.

\noindent \textbf{Loss function }
In each iteration, the Net is trained by jointly optimizing the following loss functions: hard identity classification loss $\mathcal{L}_{id}$, hard triplet loss $\mathcal L_{tri}$, soft classification loss $\mathcal L_{sid}$, soft triplet loss $\mathcal L_{stri}$. The two hard loss functions are the same as typical loss in re-ID.

\begin{equation}
\mathcal{L}_{i d}(\boldsymbol{\theta})=\frac{1}{N} \sum_{i=1}^N\mathcal{L}_{c e}\left(C\left(F\left(\boldsymbol{x}_{i} \middle| \boldsymbol{\theta}\right)\right), \tilde{\boldsymbol{y}}_{i}\right)
\end{equation}

\begin{equation}
\begin{split}
\mathcal{L}_{t r i}(\boldsymbol{\theta})=\frac{1}{N} \sum_{i=1}^{N} \max
(0,
\left\|F\left(\boldsymbol{x}_{i} \middle| \boldsymbol{\theta}\right)-F\left(\boldsymbol{x}_{i, p} \middle| \boldsymbol{\theta}\right)\right\| \\
+m-\left\|F\left(\boldsymbol{x}_{i} \middle| \boldsymbol{\theta}\right)-F\left(\boldsymbol{x}_{i, n} \middle| \boldsymbol{\theta}\right)\right\|
)
\end{split}
\end{equation}

where $\|\cdot\|$ denotes the $L^2$-norm distance, $x_i$ denotes the sampled consecutive images sampled from tracklet $i$ at the current iteration, $F\left(x_i\middle|\boldsymbol \theta\right)$ is the feature of tracklet $i$ output by current Net, $C(\cdot)$ is classifier, subscripts ${}_{i, p}$ and ${}_{i, n}$ indicate the hardest positive and hardest negative feature index in the batch, and $m = 0.5$ denotes the triplet distance margin.

The two soft loss functions use the soft pseudo labels generated by the Mean Net $E^{(T)}[\boldsymbol{\theta}]$.

\begin{equation}
\mathcal{L}_{s i d}\left(\boldsymbol{\theta}\right)=-\frac{1}{N} \sum_{i=1}^{N}\left(C(F(\boldsymbol{x}_{i} | E^{(T)}\left[\boldsymbol{\theta}\right])) \cdot \log C(F(\boldsymbol{x}_{i} | \boldsymbol{\theta}))\right)
\end{equation}

\begin{equation}
\mathcal{L}_{s t r i}\left(\boldsymbol{\theta}\right)=\frac{1}{N} \sum_{i=1}^{N} \mathcal{L}_{b c e}\left(\mathcal{T}_{i}\left(\boldsymbol{\theta}\right), \mathcal{T}_{i}\left(E^{(T)}\left[\boldsymbol{\theta}\right]\right)\right)
\end{equation}

where $\mathcal{T}_{i}$ is the soft triplet labels generated by the Mean Net. We refer the readers to \cite{ge2019mmt} for the detail definition of soft triplet loss.

The overall loss function $\mathcal L(\boldsymbol \theta)$ combines four loss functions and is formulated as,

\begin{equation}\label{eq:overall-loss}
\begin{aligned}
\mathcal{L}\left(\boldsymbol{\theta}\right) &=
\left(1-\lambda_{i d}\right) \mathcal{L}_{i d}\left(\boldsymbol{\theta}\right)
+\lambda_{i d} \mathcal{L}_{s i d}\left(\boldsymbol{\theta}\right) \\
&+\left(1-\lambda_{t r i}\right)\mathcal{L}_{t r i}\left(\boldsymbol{\theta}\right)
+\lambda_{tri} \mathcal{L}_{stri}\left(\boldsymbol{\theta}\right)
\end{aligned}
\end{equation}

where $\lambda_{id}$, $\lambda_{tri}$ are the weighting parameters.

The detailed optimization process is summarized in Algorithm~\ref{algo:main}.

\SetKwInput{KwRequire}{Require}

\begin{algorithm}
  \SetAlgoLined
\KwRequire{Ensemling momentum $\alpha$ for equation~\ref{eq:ema}, weighting factors $\lambda_{id}$, $\lambda_{tri}$ for equation~\ref{eq:overall-loss}}
  Initialize $\boldsymbol \theta$ with ImageNet pre-trained ResNet-50\;
  Intra-tracklet isolation on raw tracklets\;
  \For{$n$ in $[1, num\_epochs]$}{
    Generate hard pseudo labels from inter-tracklet association\;
    \For{each mini-batch, iteration $T$}{
        Generate soft pseudo labels from the Mean Net\;
        Update parameters $\boldsymbol\theta$ by the gradient descent of loss function equation~\ref{eq:overall-loss}\;
        Update Mean Net weights following equation~\ref{eq:ema}\;
    }
  }
  \caption{Tracklet-based Multi-level Clustering (TMC) Training Pipeline}
  \label{algo:main}
\end{algorithm}

\section{Experiments}

\subsection{Datasets}

\noindent \textbf{Video-based re-ID datasets }
Two publicly available video re-ID datasets are used in our experiments: PRID 2011~\cite{hirzer2011prid} and MARS~\cite{zheng2016mars} dataset. The MARS dataset is the largest video dataset for the person re-ID task. The dataset contains 17,503 tracklets for 1,261 identities and 3,248 distractor tracklets, which are recorded by six cameras. This dataset is split into 625 identities for training and 636 identities for testing. The PRID 2011 dataset is collected from two cameras with significant color inconsistency. It contains 385 person tracklets in camera A and 749 person tracklets in camera B. Among all persons, 200 persons are captured in both camera views. 

\noindent \textbf{Simulation }
The need for simulated tracklets stems from two aspects. Firstly, video sequences in existing video-based re-ID dataset are clean and lack of noise, which makes them unrealistic. Using MOT algorithm to generate tracklets need original video as input, however the corresponding video for video re-ID dataset is generally unavailable. Secondly, by simulating tracklets we have more control over strength of different types of noise, giving us more insight on how different noise affect re-ID model training process.

Therefore, we propose a novel algorithm to simulate tracklets from video re-ID dataset. We first formulate noise strength as ``rate of fragmentation'' ($r_{FM}$) and ``rate of switch'' ($r_{SW}$) .
Given a collection of MOT-generated tracklets
$\left\{ S_1, S_2, \dots, S_N \right\}$
and the corresponding tracking ground-truth, one can use IOU matching to find $P_{ij}$, the ground-truth person ID for the $j$-th frame in tracklet $S_i$.
Let $L_i$ be the length of tracklet $S_i$, $Q_i = \left\{ P_{ij} \mid j=1,2,\dots,L_i \right\}$ be the set of ground-truth person IDs in the tracklet. Assume the total number of ground-truth person IDs is $M$.

$r_{FM}$ is defined to be the average number of tracklets each ground-truth person ID is in.
\begin{equation}\label{eq:r_FM}
r_{F M}=\frac{1}{M} \sum_{k=1}^{M} \sum_{i=1}^{N} \mathbbm{1}_{k \in Q_{i}}
\end{equation}
$r_{SW}$ is defined to be the average number of ground-truth person IDs each tracklet contains.
\begin{equation}\label{eq:r_SW}
r_{SW} = \frac{1}{N} \sum_{i=1}^{N} \left| Q_i \right|
\end{equation}

Given $r_{FM}$, $r_{SW}$, and the number of ground-truth person IDs, the number of tracklets with noise can be calculated. The exact number of person IDs contained by each noisy tracklet are determined by observed distribution. Person IDs are randomly assigned to noisy tracklets, after which the assignment of the pure tracklets and the ``slots'' in the noisy tracklets are randomly generated. Finally, the comprising pure tracklets within each simulated tracklet are shuffled.

Our generation algorithm has the ability to generate tracklets with various noise types (\eg, tracklets with ID switch and fragmentation at the same time, tracklets with multiple ID switches). Please refer to the supplementary material for further details of the generation algorithm.

\begin{table}[]
\begin{center}
\begin{tabular}{|c|c|c|c|c|}
\hline
No. & Name & $r_{FM}$ & $r_{SW}$ & \#Tracklets \\ \hline
1 & MARS\_1.7\_1.2 & 1.7 & 1.2 & 2835  \\ \hline
2 & MARS\_2.5\_1.2 & 2.5 & 1.2 & 4023 \\ \hline
3 & MARS\_2.5\_1.5 & 2.5 & 1.5 & 3272 \\ \hline
4 & MARS & - & - & 8298 \\ \hline
\end{tabular}
\caption{Summary of generated simulation datasets}
\label{tab:sim-dataset}
\end{center}
\end{table}

Videos with static camera from the training set of MOT17~\cite{milan2016mot17} were used for estimating realistic values of $r_{FM}$ and $r_{SW}$. Two MOT algorithms, DeepSORT~\cite{wojke2017deepsort} and FairMOT~\cite{zhang2020fairmot} were picked as representatives for baseline method and recent SotA method, respectively. For realistic upper bounds of $r_{FM}$ and $r_{SW}$, the most difficult video (per person density provided by MOT17) and the less performant algorithm, DeepSORT, was used, yielding an $r_{FM}$ of value $1.7$ and $r_{SW}$ of $1.2$. For lower bounds, the easiest video (same above) and the better algorithm, FairMOT, was used, giving $r_{FM} = 2.5$, $r_{SW} = 1.2$. For the sake of comparison, we added another set of parameters $r_{FM} = 2.5$, $r_{SW} = 1.5$. Three MARS-derived datasets was generated based on these three set of $\left(r_{FM}, r_{SW}\right)$ values, summarized in Table~\ref{tab:sim-dataset}. For ease of reference, we name these datasets by their generating parameters $\left(r_{FM}, r_{SW}\right)$, \eg MARS\_1.7\_1.2.

\subsection{Evaluation Protocol}
For PRID 2011, we use the Cumulative Matching Characteristic (CMC) curve to evaluate the performance of each method. Both the averaged CMC and the mean Average Precision (mAP) are used to measure re-ID performance on MARS. Note, our method does not utilize any ID labels for model initialisation or training.

\subsection{Implementation Details}

\noindent \textbf{Intra-tracklet clustering}
In this step, we utilize DBSCAN clustering algorithm to cluster the features extracted by ResNet-50 model initialized on ImageNet. The hyper-parameters $eps$ is set to 0.6. Since ImageNet pre-trained model can already obtain good clustering results, in order to improve training efficiency, intra-tracklet isolation is only executed once before the end-to-end training process.

\noindent \textbf{Inter-tracklet clustering}
For each tracklet generated in the first step, 32 consecutive images were randomly sampled, followed by feature extraction using up-to-date model. Then we utilize DBSCAN clustering algorithm to cluster the features to generate hard pseudo labels. The hyper-parameter $eps$ is set to data-dependent. Inter-tracklet association is executed before each epoch to update the hard pseudo labels.

\noindent \textbf{End-to-end training}
ResNet-50 backbone was used, initialized by weights pre-trained on ImageNet.
The network is updated by optimizing loss with the loss weight $\lambda_{i d}=0.5$, $\lambda_{tri}=0.8$. Adam optimizer is adopted with a weight decay of 0.0005. The temporal ensemble momentum $\alpha$ is set to 0.999. The learning rate is set to 0.00035 for 40 epochs.

\subsection{Ablation Studies}

\begin{table}[]
\begin{center}
\begin{tabular}{|c|c|c|}
\hline
\multirow{2}{*}{$eps_\mathrm{inter}$} & \multicolumn{2}{c|}{MARS} \\ \cline{2-3}
& mAP & rank-1 \\ \hline
0.600 & 27.4 & 33.6 \\ \hline
0.650 & 26.2 & 32.9 \\ \hline
0.670 & 57.9 & 70.4 \\ \hline
$\mathbf{0.676^{*}}$ & \textbf{61.8} & \textbf{72.4} \\ \hline
0.680 & 57.1 & 69.2 \\ \hline
0.700 & 56.4 & 68.3 \\ \hline
\end{tabular}
\caption{Ablation studies on the value of $eps_\mathrm{inter}$ in inter-tracklet clustering. ${}^{*}$ denotes $eps_\mathrm{inter}$ value calculated by data-dependent policy.}
\label{tab:exp-inter-eps}
\end{center}
\end{table}

\noindent \textbf{Necessity of noise reduction methods }
We use intra-tracklet isolation and inter-tracklet association to reduce the ID switch and fragmentation. The necessity of inter-tracklet association has been proved in \cite{ge2019mmt}. To investigate the contribution of intra-tracklet isolation, comparative experiments are conducted on MARS and the three simulated datasets, each with and with out intra-tracklet isolation. As illustrated in Table~\ref{tab:exp-12step-on-noisy-dataset}, the experiments without intra-tracklet isolation result in much lower performances on the noisy datasets. Under the interference of ID switch, each tracklet could contain more than one person, so the hard pseudo label generated from part of the tracklet can not represent the whole tracklet. For dataset MARS without noise, the two results almost equal. In addition, the best results of the three noisy datasets are similar, proving that our method can handle varying degrees of noise.

\begin{table*}[]
\begin{center}
\begin{tabular}{|c|c|cc|cc|cc|cc|}
\hline
\multirow{2}{*}{\begin{tabular}{@{}c@{}}Intra-tracklet\\Isolation\end{tabular}} & \multirow{2}{*}{\begin{tabular}{@{}c@{}}Inter-tracklet\\Isolation\end{tabular}} & \multicolumn{2}{c|}{MARS\_1.7\_1.2} & \multicolumn{2}{c|}{MARS\_2.5\_1.2} & \multicolumn{2}{c|}{MARS\_2.5\_1.5} & \multicolumn{2}{c|}{MARS} \\ \cline{3-10} & & mAP & rank-1 & mAP & rank-1 & mAP & rank-1 & mAP & rank-1 \\ \hline
N & Y & 8.0 & 17.3 & 14.4 & 22.0 & 11.8 & 22.7 & 61.8 & 72.4 \\ \hline
Y & Y & 54.5 & 67.3 & 55.3 & 68.2 & 53.4 & 63.7 & 60.7 & 72.0 \\ \hline
\end{tabular}
\caption{Ablation studies on intra-tracklet isolation. The experiments without intra-tracklet isolation result in much lower performances on the noisy datasets. For dataset MARS without noise, the two results almost equal.}
\label{tab:exp-12step-on-noisy-dataset}
\end{center}
\end{table*}

\begin{table*}[]
\begin{center}
\begin{tabular}{|c|cccc|cccc|cccc|}
\hline
\multirow{2}{*}{$eps_\mathrm{intra}$} & \multicolumn{4}{c|}{MARS\_1.7\_1.2} & \multicolumn{4}{c|}{MARS\_2.5\_1.2} & \multicolumn{4}{c|}{MARS\_2.5\_1.5} \\ \cline{2-13} 
& noise & \#clusters & mAP & rank-1 & noise & \#clusters & mAP & rank-1 & noise & \#clusters & mAP & rank-1  \\ \hline
0.4 & 0.57 & 21720 & 35.0 & 48.8 & 1.1 & 20919 & 40.1 & 53.8 & 1.8 & 20958 & 39.1 & 52.5 \\ \hline
0.5 & 0.84 & 13847 & \textbf{54.5} & \textbf{66.2} & 1.8 & 13673 & \textbf{55.3} & \textbf{68.2} & 2.6 & 13689 & 47.1 & 57.7 \\ \hline
0.6 & 1.4 & 8963 & 50.8 & 62.0 & 2.6 & 9383 & 52.5 & 66.5 & 
3.9 & 9401 & \textbf{53.4} & \textbf{63.7} \\ \hline
0.7 & 2.5 & 6541 & 9.8 & 15.7 & 3.8 & 7352 & 15.6 & 20.1 & 7.1 & 7306 & 12.4 & 19.9 \\ \hline
\end{tabular}
\caption{Ablation studies on the value of $eps_\mathrm{intra}$. Experiments were conducted on noisy datasets. \epsinter{} is calculated by data-dependent policy for all experiments. ``Noise'' is the average noise ratio of the tracklets produces by intra-tracklet isolation. \#Clusters is the total number of tracklets produced.}
\label{tab:exp-intra-eps}
\end{center}
\end{table*}

\begin{figure}
  \begin{subfigure}[t]{.24\textwidth}
    \centering
    \includegraphics[width=\linewidth]{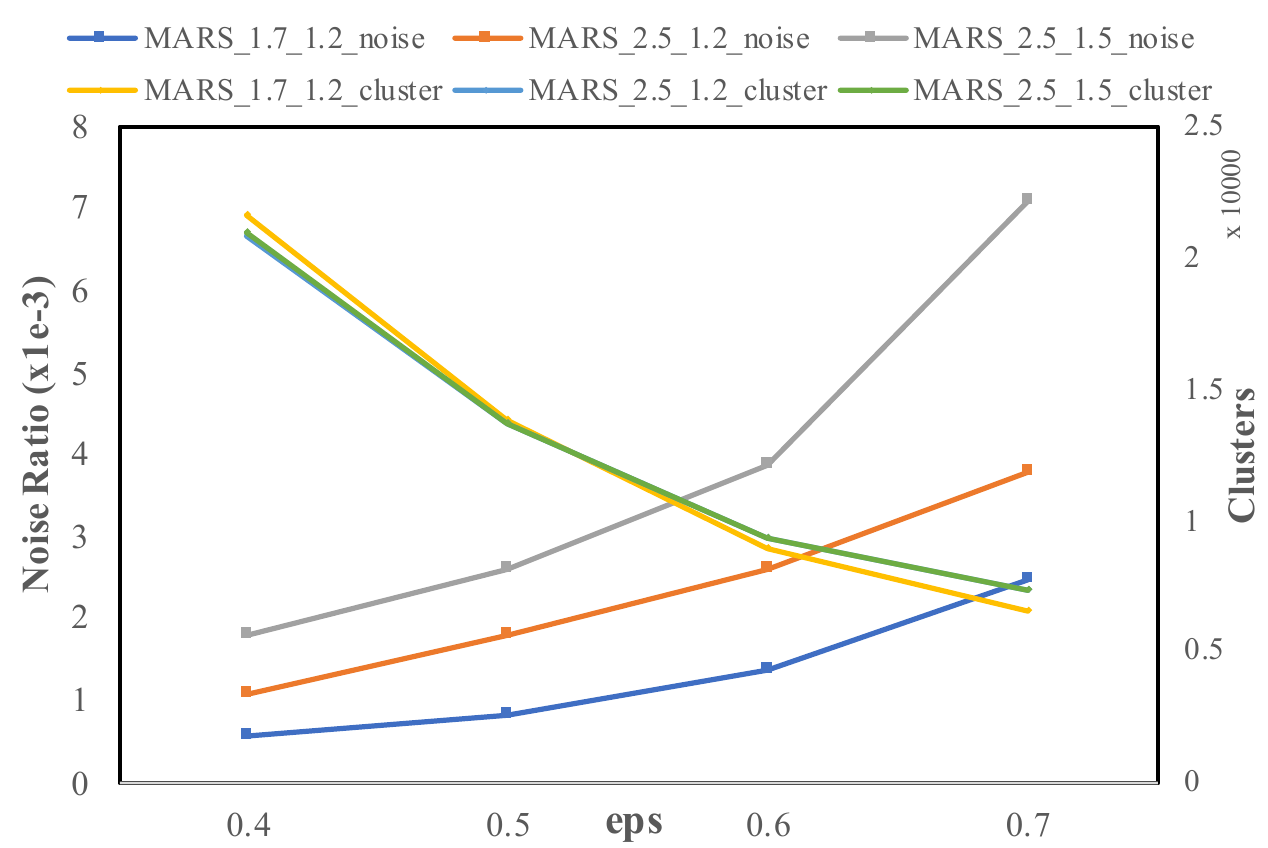}
    \caption{}
  \end{subfigure}
  \begin{subfigure}[t]{.23\textwidth}
    \centering
    \includegraphics[width=\linewidth]{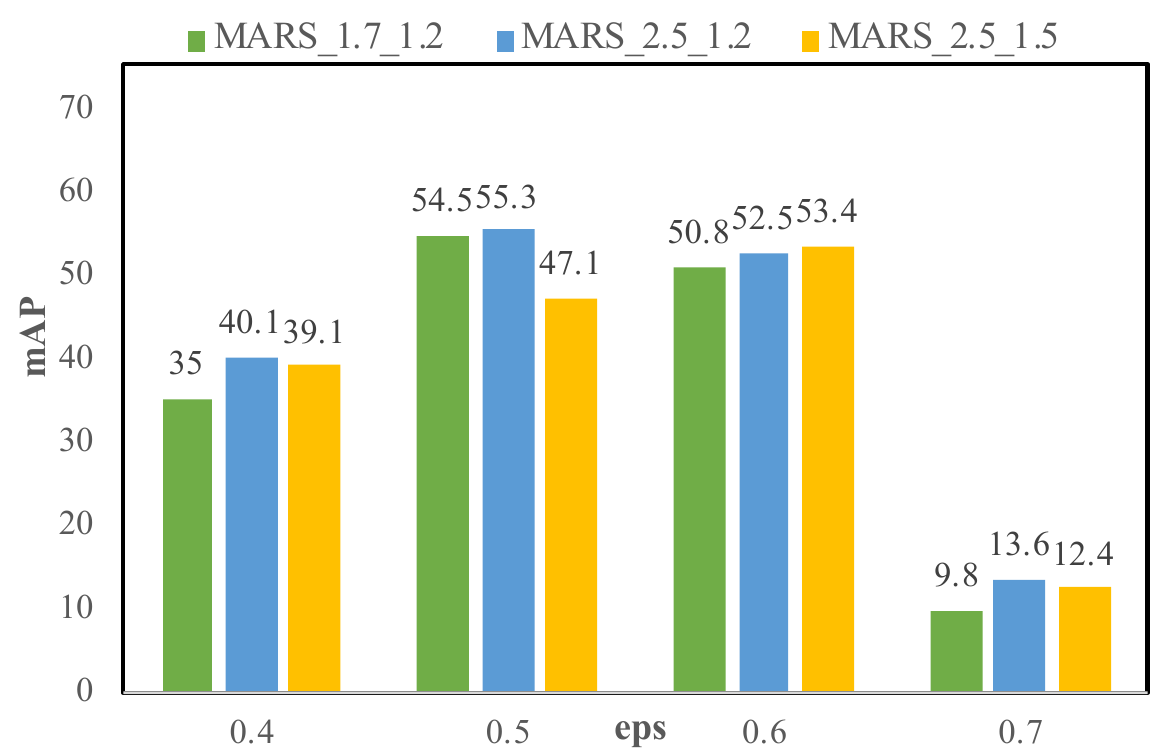}
    \caption{}
  \end{subfigure}
  
  \caption{Experiments on $eps$ value for the intra-tracklet clustering. (a) Noise ratio is larger and the number of total clusters generated is smaller with larger $eps$ value. (b) An $eps$ of value too small or too large results in less performant model learned.}
  \label{fig:intra-eps-fig}
\end{figure}

\begin{table}[]
\begin{threeparttable}
\begin{center}
\scalebox{0.8}{
\begin{tabular}{|c|ccc|cccc|}
\hline
Dataset & \multicolumn{3}{c|}{PRID 2011} &  \multicolumn{4}{c|}{MARS} \\ \hline
Rank at $r$ & 1 & 5 & 10 & 1 & 5 & \multicolumn{1}{c|}{10} & mAP \\ \hline
DAL~\cite{chen2018dal} & \textcolor{red}{\textbf{85.3}} & 97.0 & 98.8 & 46.8 & 63.9 & \multicolumn{1}{c|}{71.6} & 21.4 \\ \hline
RACE~\cite{ye2018race} & 50.6 & 79.4 & 88.6 & 41.0 & 55.6 & \multicolumn{1}{c|}{62.2} & 22.3 \\ \hline
DGM~\cite{ye2017dgm} & 61.6 & 89.0 & 94.8 &  43.8 & 60.1 & \multicolumn{1}{c|}{67.4} & 24.0 \\ \hline
DGM+~\cite{ye2019constraint} & 62.7 & 90.8 & 96.0 & 48.1 & 64.7 & \multicolumn{1}{c|}{71.1} & 29.2 \\ \hline
UTAL~\cite{li2019utal} & 54.7 & 83.1 & 96.2 & 49.9 & 66.4 & \multicolumn{1}{c|}{77.8} & 35.2 \\ \hline
EUG~\cite{wu2018eug} & - & - & - & 62.7 & 74.9 & \multicolumn{1}{c|}{82.6} & 42.5 \\ 
\hhline{|=|===|===|=|}
TMC- & 67.4 & 82.0 & 93.3 & \textcolor{red}{\textbf{72.4}} & 85.9 & \multicolumn{1}{c|}{89.3} & \textcolor{red}{\textbf{61.8}} \\ \hline
TMC & \textcolor{blue}{\textbf{68.0}} & 82.0 & 93.5 & \textcolor{blue}{\textbf{72.0}} & 84.2 & \multicolumn{1}{c|}{88.2} & \textcolor{blue}{\textbf{60.7}} \\ \hhline{|=|===|===|=|}
$\text{STAN}^\dagger$~\cite{li2018stan} & 90.3 & 98.2 & - & 82.3 & - & \multicolumn{1}{c|}{-} & 65.8 \\ \hline
$\text{SDM}^\dagger$~\cite{zhang2018sdm} & 85.2 & 97.1 & - & 71.2 & 85.7 & \multicolumn{1}{c|}{-} & - \\ \hline
\end{tabular}}
\begin{tablenotes}
\small
\item ${}^\dagger$ \text{Supervised method.}
\end{tablenotes}
\caption{Comparisons with state-of-the-art. We compare our proposed method with six unsupervised state-of-the-art methods and two supervised methods on two video re-ID datasets. ``TMC-'' is our method with intra-tracklet isolation step omitted.}
\label{tab:exp-compare-sota}
\end{center}
\end{threeparttable}
\end{table}

\noindent \textbf{Effectiveness of the clustering eps }
The two $eps$ values used in intra- and inter-tracklet clustering are closely related. We denote the values \epsintra{} and \epsinter{}, respectively. The images \epsintra{} operates on are consecutive frames of a tracklet, therefore the images are much more similar and from a small number of distinct ground-truth IDs. \epsinter{}, however, operates on all tracklets, which are less similar and from a large number of IDs.

As illustrated in Table~\ref{tab:exp-inter-eps}, the performance is sensitive to the \epsinter{} value chosen. The best performance is achieved when \epsinter{} is calculated from data distribution.

For \epsintra{} of intra-tracklet isolation, we conducted experiments on the three noisy datasets using four different \epsintra{} values. In order to ensure that different tracklets use the same clustering standard, we set the \epsintra{} to a fixed value instead of data-dependent. As illustrated in Table~\ref{tab:exp-intra-eps}, The most appropriate \epsintra{} for MARS\_1.7\_1.2 and MARS\_2.5\_1.2 is 0.5. the highest mAP value could reach 54.5\% and 55.3\%. 0.6 is the most appropriate \epsintra{} value for MARS\_2.5\_1.5, the mAP of which is 53.4\%. \textbf{1.} As shown in figure~\ref{fig:intra-eps-fig} (a), for a larger \epsintra{} value, the density requirements for a cluster to be formed are weaker. Therefore, the total number of clusters generated is smaller, and the noise ratio of clusters generated is higher. \textbf{2.} For the three simulated noisy datasets, as shown in figure~\ref{fig:intra-eps-fig} (b), if \epsintra{} is too small, the total number of clusters would be too large, presenting difficulties during the inter-tracklet association step. On the other hand, an \epsintra{} value that is too large would weaken the effect of noise reduction in intra-tracklet isolation, hindering the performance of trained model. An appropriate \epsintra{} value gives the right balance between the two effects, yielding relatively higher mAP and rank-1. \textbf{3.} \epsinter{} operates on all tracklets, which are less similar and from a large number of IDs. Moreover, it is generally preferable to have ID fragmentation noise rather than ID switch in intra-tracklet isolation result, since the former can be corrected in the subsequent inter-tracklet association step, but the latter cannot. This intuition coincides with experiment outcomes in figure~\ref{fig:intra-eps-fig} (b), where an \epsintra{} smaller than optimal value causes smaller drop in performance than an \epsintra{} larger than optimal value. We conclude that \epsintra{} should be smaller than \epsinter{}.

\subsection{Comparisons with State-of-the-Art}

We compare our proposed method with six state-of-the-art methods on two video re-ID datasets, PRID 2011 and large-scale MARS dataset. The results are shown in Table~\ref{tab:exp-compare-sota}. As for the two clean datasets without noise, our method significantly outperforms all existing video unsupervised and tracklet-based approaches on MARS, which is the dataset closest to the realistic problem. Our method even achieved competitive results against some supervised methods. Our method ranked second on PRID 2011, which we presume to be caused by the size of PRID 2011. The dataset contains relatively small number of IDs and only two tracklets are provided for each ID. This creates gap between the dataset and realistic scenarios, limiting the performance of our method. 

On MARS, the intra-tracklet isolation step in TMC might split the already-clean tracklets, having negative impact on inter-tracklet association. Therefore, results with full TMC is sightly worse than TMC with intra-tracklet isolation removed.

In addition, the method also shorten the gap in performance between supervised and unsupervised methods. Apart from this, as shown in Table~\ref{tab:exp-intra-eps}, the proposed method could achieve 55.3\% mAP, 68.2\% rank-1 on realistic tracklets with noise. Even on the tracklets with strongest noise, our method could achieve mAP 53.4\% and rank-1 63.7\%, outperforming existing unsupervised video re-ID methods even with clean tracklets. The results show that our framework outperforms existing unsupervised and weakly-supervised video re-ID methods In real-life scenarios.

\section{Conclusion}
In this work, we propose a clustering and fine-tuning framework to learn a person re-ID model from tracklets automatically generated from MOT algorithms. This is a setting that was less studied but closer to realistic problems. Our framework requires no human effort for labeling and also independent from data or model from other domains.
Characteristics of dominant noise types of MOT tracklets, i.e. ID fragmentation and ID switch, are analyzed and utilized in the noise reduction process.
Extensive experiment results on MARS and simulated tracklets of various noise show that our framework outperforms all existing unsupervised and weakly-supervised video person re-ID methods.

\section{Broader Impact}
Over the past few years, the technology of person re-ID has been deployed to benefit the society in areas such as intelligent security, smart city, and smart retail. It should be notes that, when applied maliciously, person re-ID may cause the infringement of privacy or even crime. Governments and facilities are responsible to control the usage of this technology by legislation.

{\small
\bibliographystyle{ieee_fullname}
\bibliography{egbib}
}

\end{document}